\documentclass[runningheads]{llncs}
\usepackage{graphicx}
\usepackage{amsmath}
\usepackage{amssymb}
\usepackage{booktabs}
\usepackage[svgnames,table]{xcolor}

\usepackage{pifont}
\newcommand*\CHECK{\ding{51}}
\usepackage{subcaption}
\usepackage[T1]{fontenc}
\usepackage{array, makecell}
\usepackage{listings}
\usepackage{algorithm}
\usepackage{lipsum}  
\usepackage{tikz}
\usepackage{pgfplots}
\usepgfplotslibrary{fillbetween}
\usetikzlibrary{pgfplots.groupplots}
\usepgfplotslibrary{groupplots}
\usepackage[abbreviations]{foreign}
\usepackage{enumitem}
\usepackage{multirow}
\usepackage{nicematrix}
\usepackage{mathtools}
\pgfplotsset{compat=1.18}
\usepackage{cite}
\usepackage{bbm}

\usepackage{soul}
\usepackage{xcolor}
\definecolor{codegreen}{rgb}{0,0.6,0}
\definecolor{codegray}{rgb}{0.5,0.5,0.5}
\definecolor{codepurple}{rgb}{0.58,0,0.82}
\definecolor{backcolour}{rgb}{1.0,1.0,1.0}
\lstdefinestyle{mystyle}{
	backgroundcolor=\color{backcolour},   
	commentstyle=\color{codegreen},
	keywordstyle=\color{magenta},
	numberstyle=\tiny\color{codegray},
	stringstyle=\color{codepurple},
	basicstyle=\ttfamily\scriptsize,
	breakatwhitespace=false,         
	breaklines=true,                 
	captionpos=b,                    
	numbers=left,                    
	numbersep=5pt,                  
	showspaces=false,                
	showstringspaces=false,
	showtabs=false,                  
	tabsize=1
}

\usepackage{xspace}

\usepackage[pagebackref,breaklinks,colorlinks,citecolor = blue,urlcolor=blue,linkcolor=blue]{hyperref}
\usepackage[capitalize]{cleveref}
\crefname{section}{Sec.}{Secs.}
\Crefname{section}{Section}{Sections}
\Crefname{table}{Table}{Tables}
\crefname{table}{Tab.}{Tabs.}
\usepackage{array}
\begin{document}
\title{M\&M: Tackling False Positives in Mammography with a Multi-view and Multi-instance Learning Sparse Detector}
\titlerunning{M\&M: A Multi-view and MIL Sparse Detector}
%
%
\author{Yen Nhi Truong Vu$^\star$, Dan Guo$^\star$, Ahmed Taha, Jason Su, Thomas Paul Matthews}

\authorrunning{Y.N. Truong Vu \etal}
%
\institute{WhiteRabbit.AI}
\maketitle              
\def\thefootnote{$\star$}\footnotetext{Equal Contribution}
\begin{abstract}
Deep-learning-based object detection methods show promise for improving screening mammography, but high rates of false positives can hinder their effectiveness in clinical practice. To reduce false positives, we identify three challenges: (1) unlike natural images, a malignant mammogram typically contains only one malignant finding; (2) mammography exams contain two views of each breast, and both views ought to be considered to make a correct assessment; (3) most mammograms are negative and do not contain any findings. In this work, we tackle the three aforementioned challenges by: (1) leveraging Sparse R-CNN and showing that sparse detectors are more appropriate than dense detectors for mammography; (2) including a multi-view cross-attention module to synthesize information from different views; (3) incorporating multi-instance learning (MIL) to train with unannotated images and perform breast-level classification. The resulting model, M\&M, is a \textbf{M}ulti-view and \textbf{M}ulti-instance learning system that can both localize malignant findings and provide breast-level predictions. We validate M\&M’s detection and classification performance using five mammography datasets. In addition, we demonstrate the effectiveness of each proposed component through comprehensive ablation studies.

\end{abstract}

\section{Introduction}
\label{sec:intro}
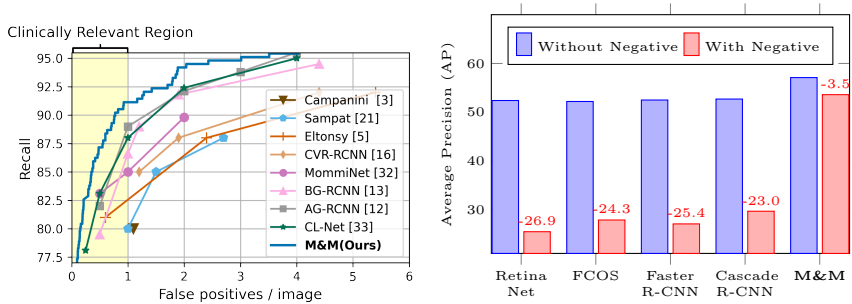
\begin{figure}[t]
    \centering
    \begin{subfigure}[b]{0.47\linewidth}
         \centering
         \includegraphics[width=\textwidth]{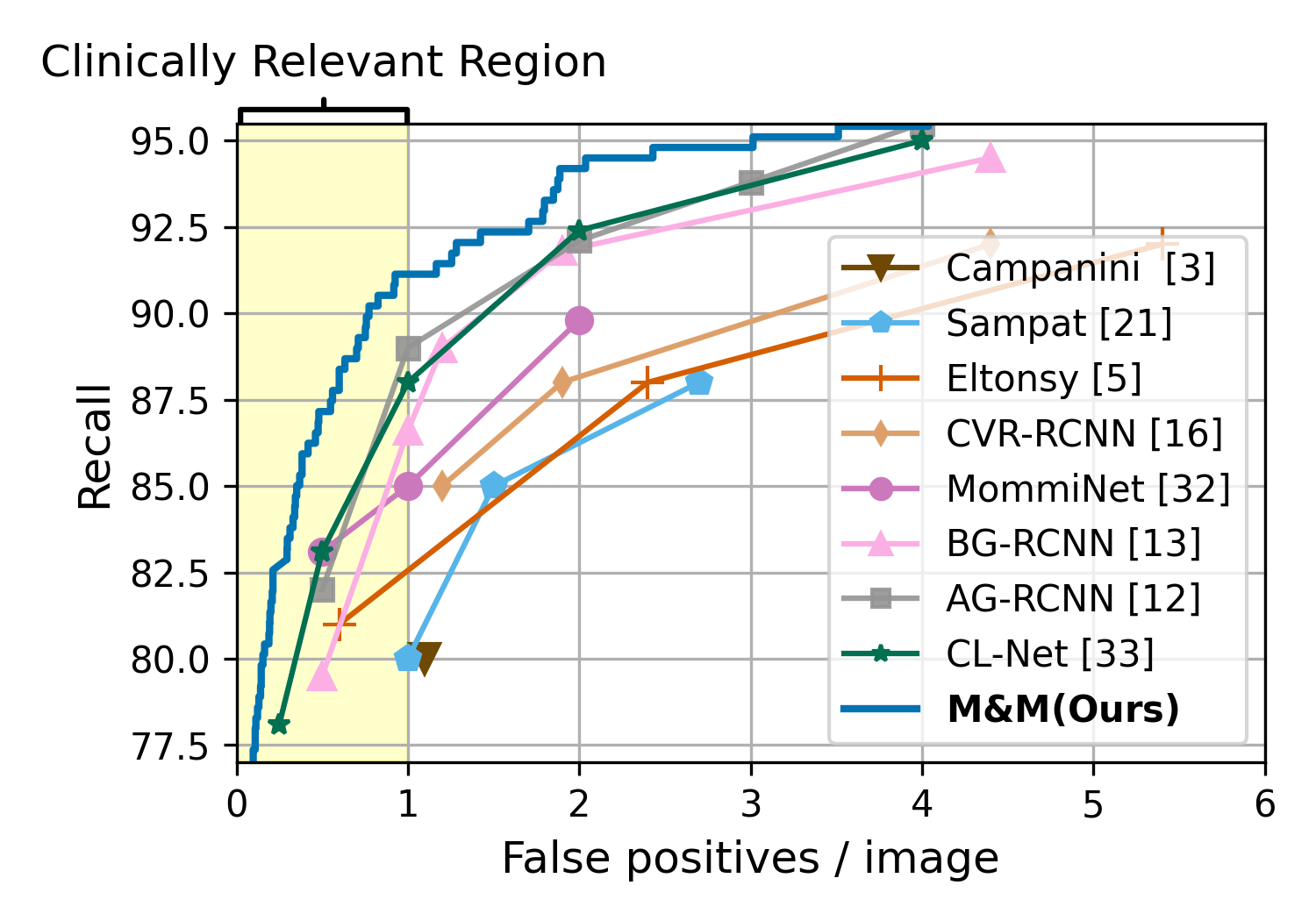}
         \caption{Free response operating characteristic (FROC) curves on DDSM.}
         \label{fig:det_sota}
    \end{subfigure}
    \hfill
    \begin{subfigure}[b]{0.52\linewidth}
    \tiny
	\begin{tikzpicture}
	\begin{axis}[ 
		ybar,
		height=0.75\textwidth,
		width=\textwidth,
		xtick=data,
		xlabel={\space},
		nodes near coords,
		nodes near coords align={vertical},
		symbolic x coords={Retina Net, FCOS, Faster R-CNN, Cascade R-CNN, M\&M},
        xticklabels={Retina Net, FCOS, Faster R-CNN, Cascade R-CNN, \textbf{M\&M}},
		ylabel=Average Precision (AP),
		ymax=70,
		ytick={30, 40, 50, 60},
		legend style={at={(0.5,0.95)}, anchor=north,legend columns=-1},
        bar width=10pt,
        legend image code/.code={
                \draw [#1] (0cm,-0.1cm) rectangle (0.2cm,0.2cm); },
        xticklabel style={align=center,text width=8mm},
		]
		\addplot+[point meta=explicit symbolic] coordinates {
			(Retina Net, 52.4) 
			(FCOS, 52.2)
			(Faster R-CNN, 52.5)
			(Cascade R-CNN, 52.7)
			(M\&M, 57.1)
		};\addlegendentry{Without Negative}
		\addplot+[point meta=explicit symbolic] coordinates {
			(Retina Net, 25.5) [\fpeval{-52.4+25.5}]
			(FCOS, 27.9) [\fpeval{-52.2+27.9}]
			({Faster R-CNN}, 27.1) [\fpeval{-52.5+27.1}]
			(Cascade R-CNN, 29.7) [-23.0]
			(M\&M, 53.6) [\fpeval{-57.1+53.6}]
		};\addlegendentry{With Negative}
	\end{axis}
\end{tikzpicture}
        \caption{Quantitative detection evaluation with and without negative images on OPTIMAM.}
         \label{fig:eval_gap}
     \end{subfigure}
    \caption{Two gaps between deep learning literature and clinical applicability. \textbf{(a)} Few works report detailed performance in the clinically relevant region of less than 1 FP/image. M\&M surpasses previous works by a large margin in this region. \textbf{(b)} Typical evaluation datasets are not representative: they contain from zero (CBIS-DDSM \cite{lee2017curated}) to few negative cases (DDSM \cite{ddsm}, INBreast \cite{moreira2012inbreast}). To illustrate the distribution shift, we train four popular dense detectors using a standard setup that includes only annotated malignant and benign cases \cite{agarwal2019automatic, cvr-rcnn, bg-rcnn}. 
    We utilize OPTIMAM \cite{optimam}, a large dataset with a significant proportion of negatives (\cref{tab:dataset}), for training and evaluation. Across all dense models, there is a large performance drop in the clinically representative setting that includes negative images. This means that the dense models are producing too many FPs on negative images. Our model, M\&M, successfully tackles this performance gap.}
\end{figure}

Screening mammography helps detect breast cancer earlier and has reduced the breast cancer mortality rate significantly \cite{duffy2020mammography}. Computer-aided diagnosis (CAD) software was developed to aid radiologists, but its effectiveness has been questioned following recent large-scale clinical studies \cite{fenton2011}. 
In particular, the high rate of false positive (FP) predictions of CAD can cause a significant reduction in radiologists’ specificity \cite{fenton2011}. Surprisingly, recent deep learning literature \cite{bg-rcnn, cvr-rcnn, campanini2004novel, sampat2008model, eltonsy2007concentric, yang2021momminet, retinamatch} focuses on improving recall without considering the need to operate at low FP rates. As shown in \cref{fig:det_sota}, most works focus on reporting recalls outside the clinically relevant region of less than 1 FP/image.

To tackle the high rate of false positives in mammography, we identify three challenges: (1) A malignant mammogram typically contains only one malignant finding. This is different from natural images: for example, an image in COCO contains on average 7.7 objects \cite{lin2014microsoft}. This calls into question the usage of dense detectors for mammography; (2) A standard screening exam consists of two views per breast. Both views are essential in making a clinical decision because a finding may appear suspicious in one view but not the other; (3) Most mammograms are negative: they do not contain any findings. However, excluding negative images from training and evaluation leads to a distribution shift since negative images are abundant in clinical practice. Concretely, the false positive rate is low for a typical evaluation data distribution but much higher for a clinically-representative data distribution, as shown in \cref{fig:eval_gap}.

In this work, we tackle these challenges and propose a \textbf{M}ulti-view and \textbf{M}ulti-instance learning system, \textbf{M\&M}. M\&M is an end-to-end system that detects malignant findings and provides breast-level classification. To achieve these goals, M\&M leverages three components: (1) Sparse R-CNN to replace dense anchors with a set of sparse proposals; (2) Multi-view cross-attention to synthesize information from two views and iteratively refine the predictions, and (3) Multi-instance learning (MIL) to include negative images during training. Ultimately, each component contributes to our goal of reducing false positives.

We validate M\&M through evaluation on five datasets: two in-house datasets, two public datasets --- DDSM \cite{ddsm} and CBIS-DDSM \cite{lee2017curated}, and OPTIMAM 
\cite{optimam}. We perform ablation studies to verify the contribution of each component of M\&M. To summarize, our contributions are:
\begin{enumerate}[noitemsep]
    \item We show that sparsity of proposals is beneficial to the analysis of mammograms, which have low disease prevalence (\cref{sec:sparse}). 
    With Sparse R-CNN, M\&M generalizes better to clinically-representative data, where the majority of images are negative, \ie, have no findings (\cref{tab:detection_evaluation});
    \item We incorporate a simple and efficient cross-view multi-head attention module for mammography analysis (\cref{sec:mv}). With multi-view reasoning, M\&M improves the recall at 0.1 FP/image by 8.6\%, as shown in \cref{tab:component_ablation};
    \item We leverage MIL to include images without bounding boxes during training (\cref{sec:mil}). Accordingly, M\&M sees seven times more images during training. With MIL, M\&M improves the recall at 0.1 FP/image by 12.6\% (\cref{tab:component_ablation}). Furthermore, M\&M can provide breast-level classification predictions, achieving AUCs of more than 0.88 on four different datasets (\cref{tab:all_cls_results}).
\end{enumerate}

\section{M\&M: A Multi-view and MIL System}
\begin{figure}[t]
    \centering
    \includegraphics[width=\linewidth]{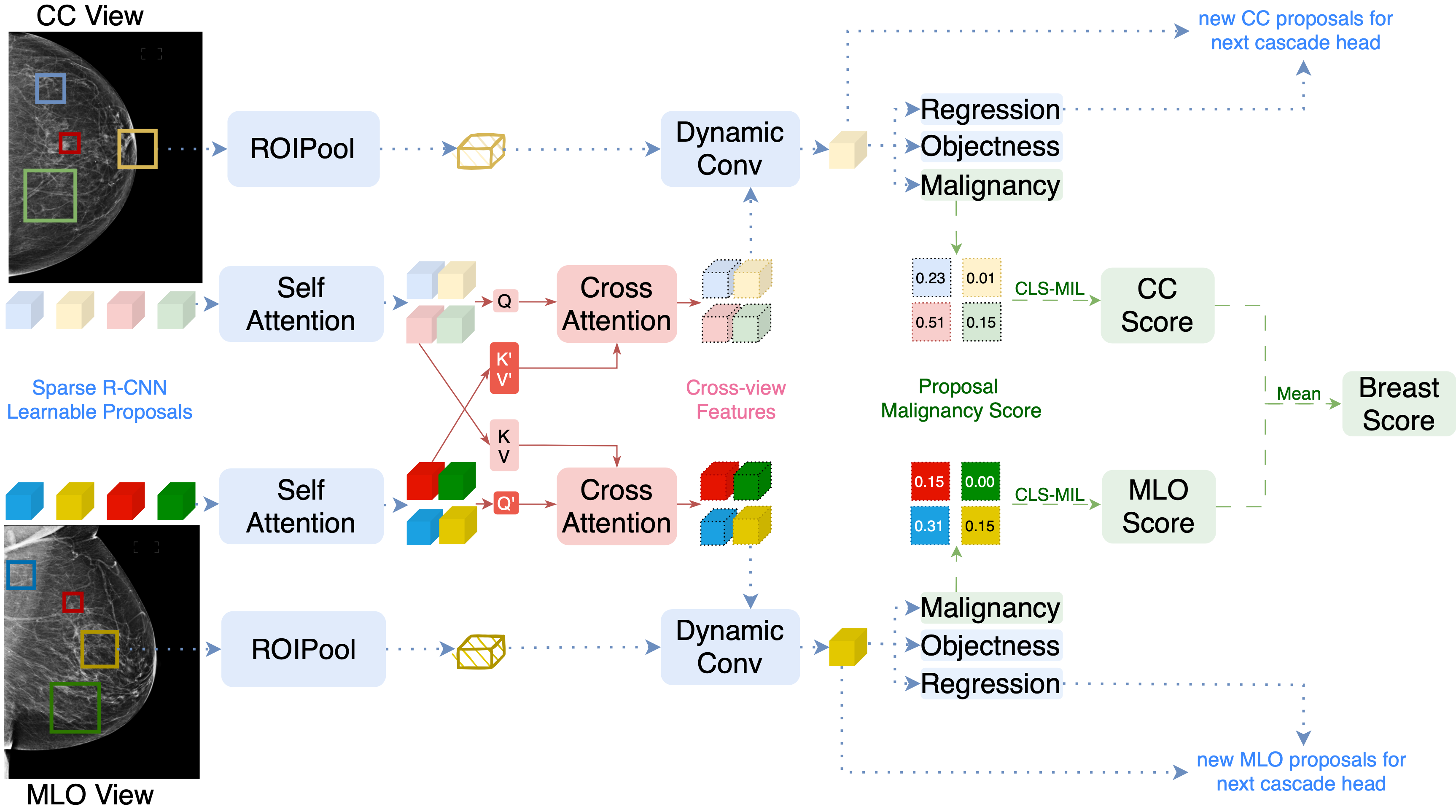}
    \caption{
    M\&M tackles false positives through (1, blue, dotted arrows) leveraging the Sparse R-CNN cascade architecture to iteratively refine sparse learnable proposals into  predictions, (2, red, solid arrows) incorporating a cross-attention module to reason about relations between objects across two views, and (3, green, dashed arrows) utilizing image and breast MIL pooling to train with images that do not have lesion annotations.
    }
    \label{fig:arch_overview}
\end{figure}
\subsection{Sparse R-CNN with Dual Classification Heads}
\label{sec:sparse}
The sparsity of malignant findings calls into question the use of dense detectors. As shown in \cref{fig:eval_gap}, dense detectors generalize poorly to negative images as they produce too many false positives. Thus, we propose to use Sparse R-CNN \cite{sparsercnn}.

Sparse R-CNN utilizes a sparse set of $N$ learnable proposals consisting of $\mathbf{b_0} \in \mathbb R^{N\times 4}$ coordinates and $\mathbf{h_0} \in \mathbb R^{N \times D}$ features. The architecture uses 6 cascading heads to iteratively refine the proposals. Within the $i^\text{th}$ head, the proposals $\mathbf{h}_{i-1}$ first interact with themselves via self-attention, and then generate DynamicConv (Fig. 4, \cite{sparsercnn}) to interact with RoI features cropped by $\mathbf{b}_{i-1}$. The resulting outputs $\mathbf{h}_i \in \mathbb R^{N \times D}$ are features for the $(i+1)^\text{th}$ head. In addition, a regression module is applied to $\mathbf{h}_i$ to generate boxes $\mathbf{b}_i \in \mathbb R^{N \times 4}$, and a classification module generates scores $\mathbf{p}_i \in \mathbb R^{N 
\times C}$, with $C$ being the number of classes.

We modify Sparse R-CNN to include dual classification modules (\cref{fig:arch_overview}). First, an objectness module produces objectness logits $\mathbf{o}_i \in \mathbb R^{N}$ to distinguish all findings --- malignant and benign --- from the background. By utilizing all findings, the objectness head increases the training sample size \cite{agarwal2019automatic, bg-rcnn, cvr-rcnn}, but also increases FPs because it flags benign findings. To mitigate this side effect, we include a dedicated malignancy module $[\mathbf{W}_i, \mathbf{b}_i]$ to generate malignancy logits $\mathbf{m}_i \in \mathbb R^{N}$ that is trained to distinguish malignant from benign findings:
\begin{align}
    \mathbf{m}_i = \mathbf{o}_i - \text{SoftPlus}(\mathbf{W}_i \mathbf{h}_i + \mathbf{b}_i). \label{eq:xmal}
\end{align}
The strictly positive function $\text{SoftPlus}(x) = \log(1+e^x)$ is chosen to enforce consistency: a high objectness logit $\mathbf{o}_i$ is required to generate a high malignancy logit $\mathbf{m}_i$. Thus, at the finding level, we obtain the following loss
\begin{equation}
\mathcal L_{\text{lesion}} =   \mathcal L_{\text{malignant}} + \mathcal L_{\text{objectness}} + 2 \mathcal L_{\text{giou}} + 5 \mathcal L_{\text{L1}}, \label{eq:lesion_loss}
\end{equation}
where $\mathcal L_{\text{giou}}$ and $\mathcal L_{\text{L1}}$ are regression losses as in Sparse R-CNN. $\mathcal L_{\text{objectness}}$ and $\mathcal L_{\text{malignancy}}$ are focal losses applied to the predicted objectness $\mathbf{o}_i$ and the predicted malignancy $\mathbf{m}_i$ across all cascading heads $1 \leq i \leq 6$, respectively.

\subsection{Multi-view Reasoning}
\label{sec:mv}
A standard screening exam includes two standard views of each breast. The craniocaudal (CC) view is taken from the top down, while the mediolateral oblique (MLO) view is captured from the side at an oblique angle. Radiologists examine both views when making a clinical decision as a finding may look innocuous in one view but suspicious in the other.

To enable multi-view reasoning, M\&M incorporates a cross-attention module \cite{vaswani2017attention} into every cascading head. Recall that within the $i^\text{th}$ cascading head, self-attention is first applied to proposal features $\mathbf{h}_{i-1}$ to reason about the relations between objects. After this self-attention module, we introduce a cross-attention module (\cref{fig:arch_overview}, Appendix Algo. 1) to reason about the relations between CC view feature $\mathbf{h}_{i-1}^{\text{CC}}$ and MLO view feature $\mathbf{h}_{i-1}^{\text{MLO}}$:
\begin{align}
    \mathbf{\tilde h}_{i-1}^{\text{CC}} & = \mathbf{h}_{i-1}^{\text{CC}} + \mathrm{MultiHeadAttn}(Q=\mathbf{h}_{i-1}^{\text{CC}},V=\mathbf{h}_{i-1}^{\text{MLO}},K=\mathbf{h}_{i-1}^{\text{MLO}}), \\
    \mathbf{\tilde h}_{i-1}^{\text{MLO}} & = \mathbf{h}_{i-1}^{\text{MLO}} + \mathrm{MultiHeadAttn}(Q=\mathbf{h}_{i-1}^{\text{MLO}},V=\mathbf{h}_{i-1}^{\text{CC}},K=\mathbf{h}_{i-1}^{\text{CC}}).
\end{align}
The enhanced embeddings $\mathbf{\tilde h}_{i-1}^{\text{CC}}$, $\mathbf{\tilde h}_{i-1}^{\text{MLO}}$ then generate DynamicConv to interact with RoI features and produce new features $\mathbf{ h}_{i}^{\text{CC}}$, $\mathbf{ h}_{i}^{\text{MLO}}$ for the $(i+1)^\text{th}$ head. Thus, with the proposed cross-attention module, the CC view's proposal features are refined iteratively using the MLO view's proposal features and vice versa.

\subsection{Multi-instance Learning}
\label{sec:mil}
Mammogram annotation is costly to obtain due to a dependency on radiologists. This high cost means that bounding boxes are often unavailable. Further, most mammograms are negative: they do not contain any findings. Yet, a model generalizes poorly if these negative images are dropped during training (\cref{fig:eval_gap}). 

Since image- and breast-level labels are available, we adopt an MIL module to include images without bounding boxes during training. To compute image- and breast-level scores, we leverage the proposal malignancy logits $\mathbf{m}_i$ (\cref{eq:xmal}). Since an image is malignant if it contains a malignant lesion, we obtain image-level scores by applying the NoisyOR function $f(\mathbf x) = 1 - \prod_{k=1}^N (1-\mathbf{x}[k])$ to the malignancy probabilities $\mathbf{p}_i = \mathrm{Sigmoid}(\mathbf{m}_i) \in \mathbb R^N$. Next, as CC and MLO views offer complimentary information on a breast, we obtain breast-level malignancy score by averaging the image-level scores across these views.

We apply cross-entropy losses $\mathcal L_{\text{image}}$ and $\mathcal L_{\text{breast}}$ at the image and breast level for all training samples. The lesion loss $\mathcal L_{\text{lesion}}$ (Eq. \eqref{eq:lesion_loss}) is only applied for annotated lesions. We thus obtain the following total training loss for M\&M:
\begin{equation}
\mathcal L = \mathbbm{1}_{\text{annotated lesion}}\mathcal L_{\text{lesion}} + 0.5\mathcal L_{\text{image}} + 0.5\mathcal L_{\text{breast}}.
\end{equation}

\section{Experiments}

\subsubsection{\underline{Implementation Details.}}
We use PyTorch 1.10. The training settings follow Sparse R-CNN \cite{sparsercnn}. We apply random horizontal flipping and random rotation. We resize the images' shorter edges to 2560 with the larger edges no longer than 3328. We utilize a COCO-pretrained PVT-B2-Li backbone \cite{wang2022pvt}. 
We use AdamW optimizer 
with $5 \times 10^{-5}$ learning rate and $0.0001$ weight decay. The model is trained for 9000 iterations, and the learning rate is scaled by 0.1 at the 6750 and 8250 iterations. Each batch contains 16 breasts (32 images). We employ a 1:1 sampling ratio between unannotated and annotated images.

\begin{table}[t]
    \centering
    \scriptsize
    \setlength{\tabcolsep}{7pt}
    \caption{Dataset statistics. We report the number of breasts in each dataset, broken down by 3 categories: malignant, benign, and negative. Malignant breasts contain findings with positive biopsy outcomes. Benign breasts contain findings that are determined to be non-malignant after additional follow-up. Negative breasts do not contain any radiologist-marked findings. In the parentheses, we report the number of breasts with bounding box annotations. ``Bbox'' indicates whether bounding box annotations are available.}
    \begin{NiceTabularX}{\linewidth}{@{}lcr@{\hspace{2pt}}rr@{\hspace{2pt}}rr@{\hspace{2pt}}r@{}}
    \CodeBefore
            \rowcolors{2}{}{lightgray!20}
        \Body
        \toprule
        Datasets & Bbox  & Malignant & (Ann.) & Benign & (Ann.) & Negative & (Ann.) \\ 
        \midrule
        OPTIMAM & \CHECK & 4,838 & (4,245) & 1,999 & (567) & 26,003 & (2)\\
        Inhouse-A & & 496 & (0) & 2,128 & (0) & 2074 & (0)\\
        Inhouse-B & & 243 & (0) & 7,797 & (0) & 47,929 & (0)\\
        DDSM & \CHECK & 624 & (624) & 555 & (555) & 2,877 & (1) \\
        CBIS-DDSM & \CHECK  & 312 & (310) & 347 & (336) & 0 & (0)\\
        \bottomrule
    \end{NiceTabularX}
    \label{tab:dataset}
\end{table}

\subsubsection{\underline{Datasets.}} We utilize three 2D digital mammography datasets: (1) \textit{OPTIMAM}: a development dataset derived from the OPTIMAM database \cite{optimam}, which is funded by Cancer Research UK. We split the data into train/val/test with an 80:10:10 ratio at the patient level; (2) \textit{Inhouse-A}: an evaluation dataset collected from a U.S. multi-site mammography operator; (3) \textit{Inhouse-B}: an evaluation dataset collected from a U.S. academic hospital (see \cite{pedemonte2022deep}, Sec. 2.2 for more details on the inhouse datasets). We also utilize two film mammography datasets: (4) \textit{\href{http://www.eng.usf.edu/cvprg/mammography/database.html}{DDSM}:} a dataset maintained at the University of South Florida \cite{ddsm}. We followed the methods by \cite{campanini2004novel, eltonsy2007concentric, cvr-rcnn, bg-rcnn} to split the test set; (5) \textit{\href{https://wiki.cancerimagingarchive.net/pages/viewpage.action?pageId=22516629\#225166298fe72482d5b94b979faa31a2a90dad3f}{CBIS-DDSM}:} a curated subset of DDSM \cite{lee2017curated}. We only include breasts that have one CC view and one MLO view. Dataset statistics are reported in \cref{tab:dataset}.

\subsubsection{\underline{Metrics.}} We report average precision with Intersection over Union from 0.25 to 0.75. $\text{AP}_{\text{mb}}$ denotes average precision on the set of annotated malignant and benign images. $\text{AP}$ denotes average precision when all data is included. We report free response operating characteristic (FROC) curves and recalls at various FP/image (R@t). Following  \cite{campanini2004novel,eltonsy2007concentric, cvr-rcnn, vu2022wrdet}, a proposal is considered true positive if its center lies within the ground truth box. For classification, we report the area under the receiver operating characteristic curve (AUC).

\begin{table}[t]
\setlength{\tabcolsep}{7pt}
\caption{Quantitative detection evaluation on OPTIMAM. $\Delta$ denotes the AP gap between evaluating with and without negative images.}
\label{tab:detection_evaluation}
\centering
\scriptsize
\begin{tabular}{@{}ll*6r@{}}
\toprule
Model & $\text{AP}_{\text{mb}}$ & $\text{AP}$ & $\Delta$ & R@0.1 & R@0.25 & R@0.5 \\
\midrule
RetinaNet \cite{lin2017focal}  & 52.4 & 25.5  & -26.9 & 53.3 & 73.1 & 83.0 \\
FCOS \cite{tian2019fcos} & 52.2 & 27.9 & -24.3 & 52.0 & 77.4 & 87.0 \\
Faster R-CNN \cite{ren2015faster} & 52.5 &  27.1 & -25.4  & 51.5 & 71.2 & 84.1 \\
Cascade R-CNN \cite{cai2018cascade} & 52.7 &  29.7 & -23.0  & 54.9 & 77.0 & 86.2 \\
Sparse R-CNN \cite{sparsercnn} & 53.2  &  36.2 & -17.0  & 64.3 & 77.0 & 85.5  \\
\textbf{M\&M (ours)}  & \textbf{57.1}  & \textbf{53.6}   & \textbf{-3.5} & \textbf{87.7}      &  \textbf{90.9}      &  \textbf{92.5}    \\
\bottomrule
\end{tabular}
\end{table}

\begin{figure}[t]
\centering
\begin{subfigure}[b]{0.47\linewidth}
\includegraphics[width=\linewidth]{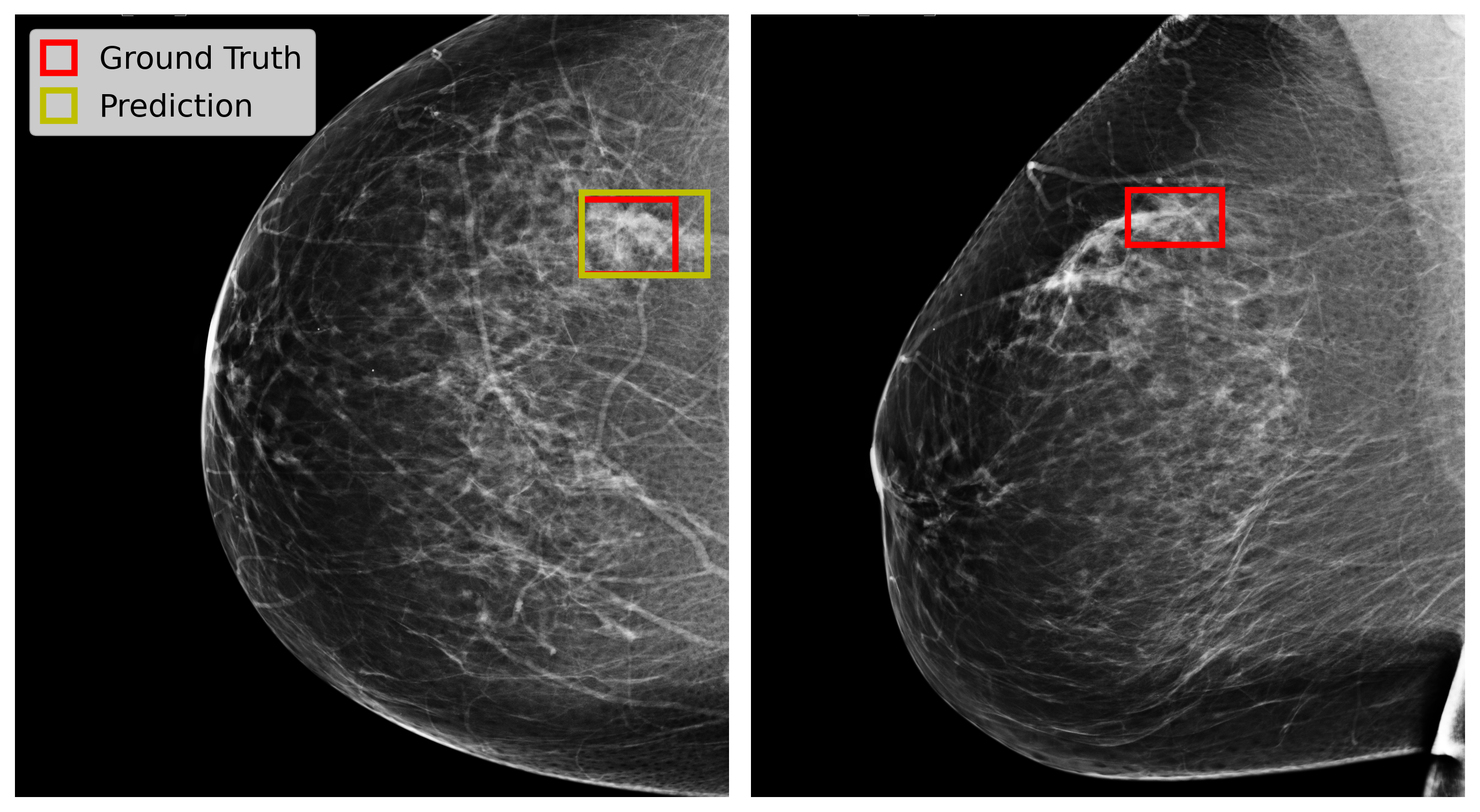}
\end{subfigure} \hfill
\begin{subfigure}[b]{0.47\linewidth}\includegraphics[width=\linewidth]{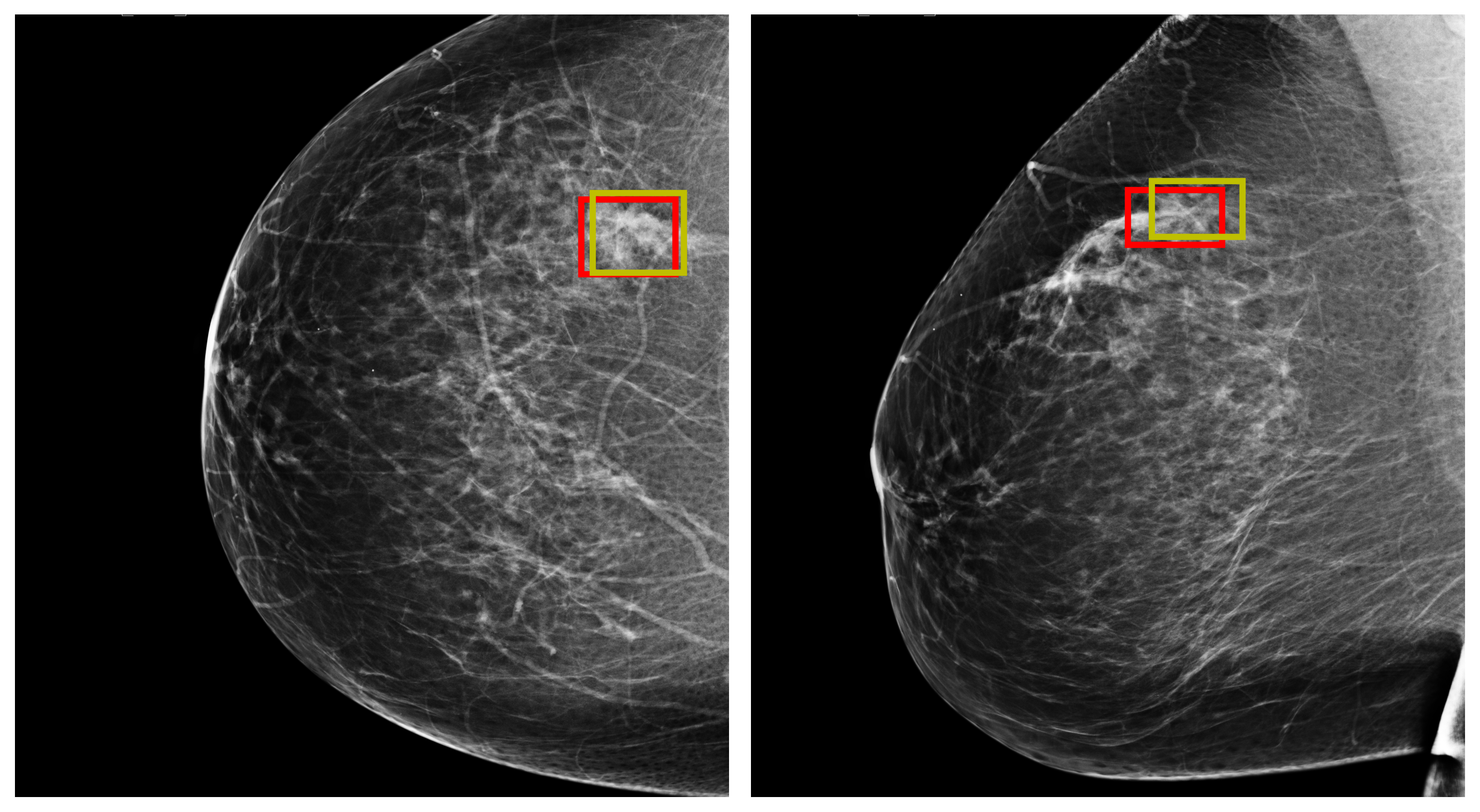}
\end{subfigure} 
\caption{Qualitative Evaluation. \textbf{Left}: Model without multi-view (row 4 of \cref{tab:component_ablation}) produces a loose box on the CC view and misses the finding on the MLO view. \textbf{Right}: M\&M produces tight boxes around the finding in both views.}
\label{fig:mv_qualitative}
\end{figure}

\subsubsection{\underline{Detection Results.}}
\cref{tab:detection_evaluation} presents quantitative detection evaluation on OPTIMAM.
All dense detectors \cite{lin2017focal, tian2019fcos, ren2015faster, cai2018cascade} 
suffer a large $\Delta$ gap of more than 23 points (pt) between excluding and including negative images. Large $\Delta$ means the models are producing too many FPs on negative images. Sparse R-CNN \cite{sparsercnn} generalizes significantly better with a gap of 17pt. This shows the importance of sparsity for reducing FP. By adding both multi-view and MIL, M\&M successfully reduces the $\Delta$ gap to 3.5pt. With this performance gap closed, M\&M is able to achieve a high recall of 87.7\% at just 0.1 FP/image.

\cref{fig:det_sota} compares M\&M with recent literature evaluated on DDSM. M\&M adopts the same DDSM splits used by \cite{campanini2004novel, bg-rcnn, cvr-rcnn, zhao2022check, liu2021act}, while \cite{eltonsy2007concentric, sampat2008model, yang2021momminet} use other splits. M\&M (87\% R@0.5) outperforms all recent SOTA with the same test split, including 2022 SOTA \cite{zhao2022check} (83\% R@0.5), by at least 4\%.

\begin{table}[t]
\centering
\caption{Quantitative classification evaluation. (a) On three private datasets, we use two open-sourced mammography classifiers as baselines \cite{shen2021interpretable, taha2022deep}. All models were trained only on OPTIMAM. We report AUC at both the breast and the exam level, except for Inhouse-A, where breast-level labels are unavailable. (b) We train M\&M on CBIS-DDSM and compare breast AUC with recent literature. (* Tulder \etal \cite{tulder2021multi} report results using five-fold cross validation.)} \label{tab:all_cls_results}
\begin{subtable}[t]{0.5\linewidth}
\caption{Private Datasets}
\vspace{3pt}
\centering
\scriptsize
\begin{tabular}{@{\extracolsep{\fill}}lccccc@{}}
\toprule
     \multirow{2}{*}{Model} & \multicolumn{2}{c}{OPTIMAM} & Inhouse-A & \multicolumn{2}{c}{Inhouse-B} \\
     \cmidrule(lr){2-3} \cmidrule(lr){4-4} \cmidrule(lr){5-6}
     & \makecell[l]{Breast\\AUC}   & \makecell[l]{Exam\\AUC}    & \makecell[l]{Exam\\AUC}  & \makecell[l]{Breast\\AUC}     & \makecell[l]{Exam\\AUC}     \\
     \midrule
GMIC \cite{shen2021interpretable} & 0.911 & 0.896 & 0.814 & 0.815 & 0.796 \\
HCT \cite{taha2022deep}  & 0.923 & 0.912 & 0.816 & 0.817 & 0.793 \\
\textbf{M\&M (ours)} & \textbf{0.960} & \textbf{0.942} & \textbf{0.920} & \textbf{0.910} & \textbf{0.898} \\
\bottomrule
\end{tabular}
\label{tab:classification_evaluation}
\end{subtable}
\hfill
\begin{subtable}[t]{0.4\linewidth}
\caption{CBIS-DDSM}
\centering
\scriptsize
    \begin{tabular}{@{}lc@{}}
        \toprule
        Model & \makecell{Breast\\AUC} \\
        \midrule
        ResNet50 \cite{lopez2022multi} & 0.724\\
        Shared ResNet \cite{wu2020improving} & 0.735  \\
        PHResNet50 \cite{lopez2022multi} & 0.739 \\
        Cross-view Transformer \cite{tulder2021multi} & \, 0.803$^{*}$ \\
        \textbf{M\&M (ours)} & \textbf{0.883} \\
        \bottomrule
    \end{tabular}
    \label{tab:class_sota}
\end{subtable}
\end{table}

\subsubsection{\underline{Classification Results.}}  
\cref{tab:classification_evaluation} reports M\&M's breast-level and exam-level classification results on OPTIMAM and the two inhouse datasets. We use GMIC \cite{shen2021interpretable} and HCT \cite{taha2022deep} as baselines since they are open-sourced classifiers developed  for mammography. All three models were trained only on OPTIMAM. For all models, the breast-level score is the average of the CC score and MLO score, while the exam-level score is the max of the left breast score and right breast score. Both baseline models suffer large generalization drops of approximately 0.08--0.12 exam AUC when evaluated on Inhouse-A and Inhouse-B. In comparison, M\&M has smaller performance gaps of 0.02 on Inhouse-A and 0.04 on Inhouse-B. Similar observations for other classifiers, such as EfficientNet, are reported in the appendix.

\cref{tab:class_sota} compares M\&M with recent literature reporting on the public CBIS-DDSM dataset. In particular, M\&M outperforms the cross-view transformer \cite{tulder2021multi} and PHResNet50 \cite{lopez2022multi} by 0.08 and 0.14 breast AUC, respectively.

\subsubsection{\underline{Qualitative Evaluation.}}
\cref{fig:mv_qualitative} presents a qualitative evaluation of the multi-view module. With multi-view, M\&M produces a tighter box on the CC view and recovers a missed finding on the MLO view.

\subsubsection{\underline{Ablation Studies.}}
\label{sec:ablation}
\begin{figure}[t]
\noindent
\begin{minipage}[t]{0.5\linewidth}
\vspace{5pt}
\scriptsize
\renewcommand{\arraystretch}{1.33}
\begin{NiceTabular}[t]{cccrrrrc}
    \CodeBefore
        \rowcolors{3}{}{lightgray!20}
    \Body
    \toprule
    Dual & Multi  & MIL & $\text{AP}_{\text{mb}}$  & $\text{AP}$  & $\Delta$ & R@0.1 & Breast \\
    heads & -view & & & & & & AUC\\
    \midrule
    & & & 50.7 & 35.4 & -15.3 & 65.1 & - \\ 
    \CHECK & & & 54.1 & 40.4 & -13.7 & 67.8 & - \\
    \CHECK & \CHECK & & 54.3 & 48.4 & -5.9 & 76.4 & - \\
    \CHECK &  & \CHECK & 54.5 & 50.9 & -3.6 & 80.4 & 0.950 \\
     \CHECK & \CHECK & \CHECK & \textbf{55.2} &\textbf{55.6} & \textbf{0.4} & \textbf{86.3} & \textbf{0.954} \\
    \bottomrule
\end{NiceTabular}
\end{minipage} \hfill
\begin{minipage}[t]{0.5\linewidth}
\vspace{0sp}
    \includegraphics[width=\linewidth]{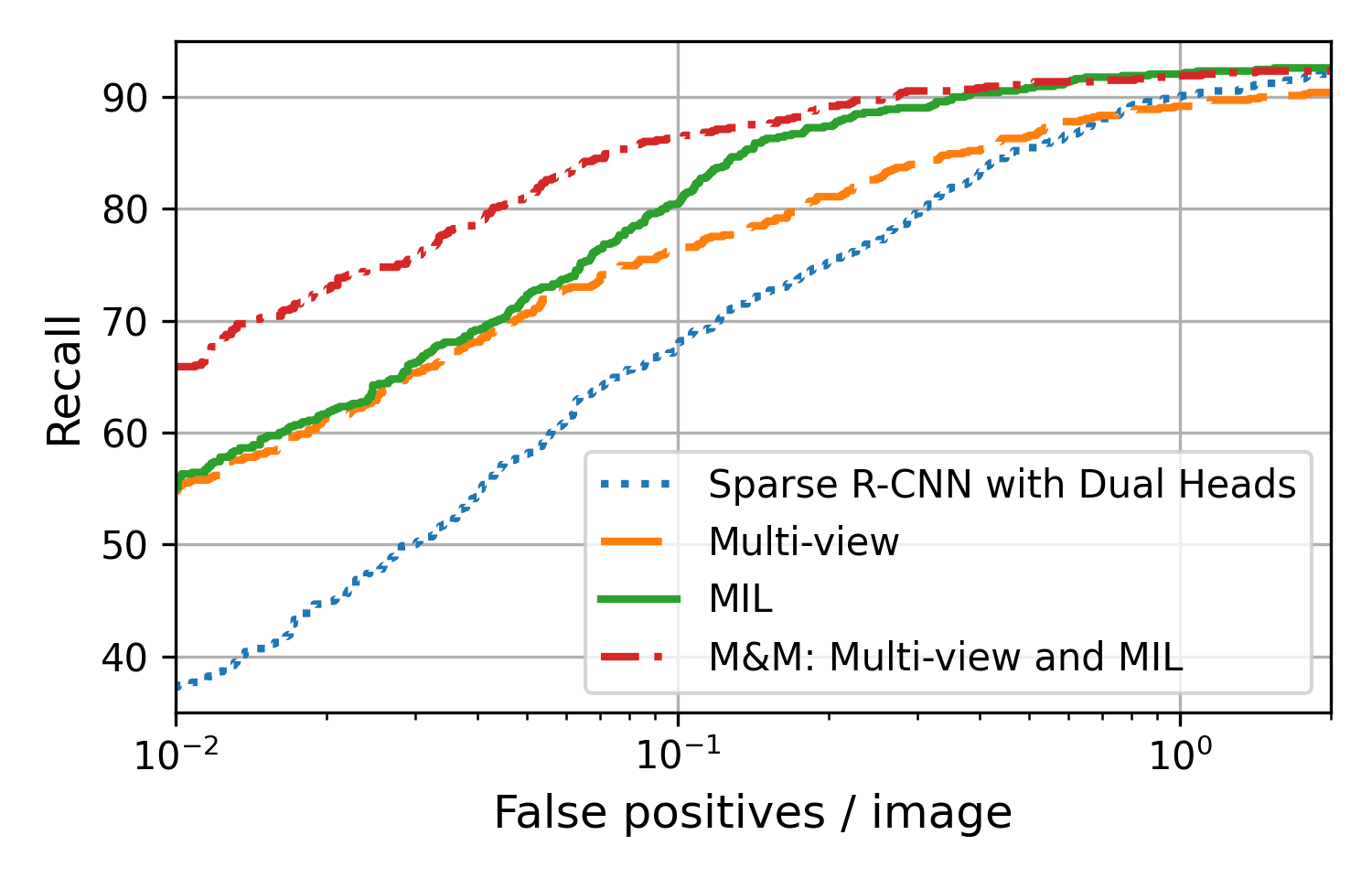}
\end{minipage}
\caption{Effect of M\&M's components on classification and detection performance.}
\label{tab:component_ablation}
\end{figure}

\cref{tab:component_ablation} presents ablation results using the OPTIMAM validation split. On the left, we demonstrate how each component of M\&M contributes to closing the gap $\Delta$ between evaluating with and without negative images. Notably, without using any extra training samples, multi-view reasoning reduces $\Delta$ to only $-5.9$pt (Row 3). MIL allows the model to train with significantly more negative images, reducing $\Delta$ to $-3.6$pt (Row 4). On the right of \cref{tab:component_ablation}, the FROC curves show how each component of M\&M improves recall significantly at low FP/image. In particular, M\&M's recall at 0.1FP/image is $86.3\%$, $+21.2\%$ over vanilla Sparse R-CNN.

\subsubsection{\underline{Further studies.}} In the appendix, we present more qualitative evaluation as well as further ablation studies on (1) number of learnable proposals, (2) different MIL schemes, (3) backbone choices and (4) positional encoding.

\noindent  

\section{Discussion and Conclusion}
We present M\&M, an end-to-end model leveraging multi-view reasoning and multi-instance learning for mammography detection and classification. 

As a detector, M\&M offers significant improvement in recall at low FP/image (\cref{fig:det_sota}, \cref{tab:detection_evaluation}). This success comes from three points of advancement. First, unlike previous works that do not consider the impact of sparsity \cite{cvr-rcnn, bg-rcnn, zhao2022check}, we show that sparsity of proposals is beneficial for false positive reduction (\cref{tab:detection_evaluation}). Second, M\&M incorporates multi-view reasoning through iterative application of cross-attention and proposal refinement in the cascading heads. M\&M's multi-view module is effective (\cref{tab:component_ablation}) yet simple, requiring neither positional encoding \cite{cvr-rcnn, yang2021momminet, bg-rcnn} nor extra proposal correspondence annotations \cite{zhao2022check}. Finally, our MIL formulation allows for training with representative data distribution in an end-to-end one stage pipeline. This is more advantageous than previous pipelines that require additional stages or classifiers to reduce false positives \cite{deephealth, sarath2020two, vu2022wrdet}.

As a classifier, M\&M establishes strong performance on several datasets (\cref{tab:all_cls_results}). M\&M offers two advantages over image classifiers: (1) Image classifiers are often pre-trained as patch classifiers with patches cropped from bounding box annotations \cite{shen2021interpretable, lopez2022multi, taha2022deep}. In comparison, M\&M utilizes these bounding boxes to learn localization and can be trained directly in a single stage from COCO/ImageNet weights; (2)~Image classifiers offer limited explainability, while M\&M's breast-level prediction is more interpretable through its localization ability.

{\small
\bibliographystyle{splncs04}
\bibliography{paper3448}
}

%
%
%
%

\newpage
\appendix
\setcounter{page}{1}
\renewcommand{\thefigure}{A\arabic{figure}}
\renewcommand{\thetable}{A\arabic{table}}
\setcounter{table}{0}
\setcounter{figure}{0}

\begin{algorithm}[H]
    \tiny
	\caption{M\&M Multi-view Cross Attention. This module is to be called on L191 in the official implementation of \href{https://github.com/PeizeSun/SparseR-CNN/blob/dff4c43a9526a6d0d2480abc833e78a7c29ddb1a/projects/SparseRCNN/sparsercnn/head.py\#L191}{Sparse R-CNN head}.}\label{alg:cross_attention}
    \tiny
    \begin{tabular}{@{}c@{}}
    \begin{lstlisting}[language=Python]
class MultiviewCrossAttn(nn.Module):
  def __init__(self, hdim=128, nhead=8, dropout=0.1):
    self.mv_atn = nn.MultiheadAttention(hdim, nhead)
    self.dropout_mv = nn.Dropout(dropout)
    self.norm_mv = nn.LayerNorm(hdim)
  def forward(self, pro_features):
    # collect CC and MLO features from the batch dimension
    cc_feats = pro_features[:, :pro_features.shape[1]//2]
    mlo_feats = pro_features[:, pro_features.shape[1]//2:]
    # cross attention to enhance CC features
    cc_feats2,_ = self.mv_atn(query=cc_feats, key=mlo_feats, value=mlo_feats)
    cc_feats += self.dropout_mv(cc_feats2)
    cc_feats = self.norm_mv(cc_feats)
    [...] # vice versa for MLO, omitted here due to space
    pro_features = torch.stack(cc_feats, mlo_feats, dim=1) # restacking
    return pro_features # new features used for inst_interact (DynamicConv)
    \end{lstlisting}
    \end{tabular}
\end{algorithm}
\vspace{-1cm}
\begin{table}[H]
    \centering
    \caption{Ablation study on (a) effect of number of learnable proposals in a multi-view only model, and (b) effect of positional encoding. Results are on OPTIMAM validation set.}
    \begin{subtable}[t]{0.48\linewidth}
    \caption{The gap between evaluating with and without negatives $\Delta$ worsens as $N$ increases, showing that multi-view reasoning benefits from sparsity.}
    \scriptsize
    \begin{tabular}{@{}*5c@{}}
        \toprule
        No. Proposals & Training Time & $\text{AP}_{\text{mb}}$ & AP & $\Delta$ \\
        \midrule
        10 & 7.5h & 53.6 & 48.4 & \textbf{-5.2}\\
        40 & 7.7h & \textbf{54.3} & \textbf{48.4} & -5.9\\
        100 & 8.0h & 53.2 & 47.1 & -6.1 \\
        400 & 8.9h & 53.0 & 45.8 & -7.2 \\
        \bottomrule
    \end{tabular}
    \label{tab:num_proposal}
    \end{subtable} \hfill
    \begin{subtable}[t]{0.48\linewidth}
    \caption{Different from \cite{cvr-rcnn, bg-rcnn, zhao2022check}, we found that positional encodings deliver insignificant boosts in AP and breast AUC. Our observation is similar to Sparse R-CNN's observations (~\cite{sparsercnn}, Tab. 10).}
    \scriptsize
    \begin{tabular}{lccc}
        \toprule
        Positional Encoding & $\text{AP}_{\text{mb}}$ & $\text{AP}$ & Breast AUC
        \\
        \midrule
        None & \textbf{55.2} & 55.6 & 0.954 \\
        Proposal center & 55.1 & 55.2 & \textbf{0.955} \\
        Nipple distance & 54.4 & \textbf{55.7} & \textbf{0.955}\\
        \bottomrule
    \end{tabular}
    \label{tab:position_choice}
    \end{subtable}
\end{table}
\vspace{-1cm}
\begin{table}[H]
\setlength{\tabcolsep}{5pt}
    \centering
    \caption{Effect of MIL approach. We also experiment with learnable image-MIL by applying a FC layer on (1) GAP: the Global Average Pooled proposal features, and (2) CLS-token: a BERT-like token that summarizes proposal features. Results are on OPTIMAM validation set.}
    \scriptsize
    \begin{NiceTabularX}{\linewidth}{@{}p{2.15cm}p{2.15cm}*3cc@{}}
        \toprule
        \makecell[l]{Image MIL} & \makecell[l]{Breast MIL} & $\text{AP}_{\text{mb}}$ & $\text{AP}$ & $\Delta$ & \makecell[l]{Breast \\ AUC}
        \\
        \midrule
        \multirow{2}{*}{Max} & Max &54.7  &54.0 &-0.7 &0.952  \\
         & Mean &54.5 &54.6 & 0.1 & 0.953 \\
        \midrule
        \multirow{2}{*}{Noisy-OR} & Max &55.1 &55.5 & \textbf{0.4} & {\bf 0.954} \\
         & Mean &55.2 &{\bf 55.6} & \textbf{0.4} & {\bf 0.954}\\
         \midrule
        \multirow{2}{*}{GAP} & Max &55.9 & 54.6 &-1.3 & 0.954 \\
         & Mean  &56.1 &54.0 & -2.1 &0.949 \\
        \midrule
        \multirow{2}{*}{CLS-token} & Max &55.3  &53.7&-1.6  &0.947 \\
         & Mean & 56.2 &55.0 & -1.2 &0.951 \\
        \bottomrule
    \end{NiceTabularX}
    \label{tab:more_mil_choice}
\end{table}

\begin{table}[h!]
\setlength{\tabcolsep}{2pt}
\caption{Quantitative detection evaluation with different backbones on two test sets. On OPTIMAM, across all different backbones, M\&M has a small gap $\Delta$ between evaluating with and without negative images. On DDSM, M\&M achieves more than 83\% recall at 0.5 FP/image across three different backbones.}
\label{tab:backbone_det}
\centering
\scriptsize
\begin{tabular*}{\linewidth}{@{\extracolsep{\fill}}p{1.4cm}l*6r@{}}
\toprule
Dataset & Backbone & $\text{AP}_{\text{mb}}$ & $\text{AP}$ & $\Delta$ & R@0.1 & R@0.25 & R@0.5 \\
\midrule
\multirow{4}{*}{OPTIMAM} & GMIC & 50.7 & 44.6  & -6.1 & 75.4  & 82.5 & 86.5 \\
& EfficientNet-B0 & 51.9 & 45.4  &  -6.5  & 78.8  & 87.1 & 89.7 \\
& ResNet-50 & 52.8 & 47.0 &  -5.8  & 80.4  & 87.2 & 90.0 \\
& PVT-B2-Li  & \textbf{57.1}  & \textbf{53.6}   & \textbf{-3.5} & \textbf{87.7}      &  \textbf{90.9}      &  \textbf{92.5}     \\
\midrule
\multirow{4}{*}{DDSM} & GMIC & 31.1 & 27.0 & -4.1 & 52.9  & 72.2 & 79.2 \\
& EfficientNet-B0 & 39.1 & 35.2 & -3.9 & 66.4 & 74.6 & 83.2 \\
& ResNet-50 & 38.9 & 36.6  &  -2.3  & 75.1  & 79.2 & 83.5 \\
& PVT-B2-Li & \textbf{39.2} & \textbf{37.0} & \textbf{-2.2} & \textbf{80.4} & \textbf{82.6} & \textbf{87.2}  \\
\bottomrule                         
\end{tabular*}
\end{table}
\vspace{-0.5cm}
\begin{figure}[H]
\centering
\begin{subfigure}[b]{0.47\linewidth}
\includegraphics[width=\linewidth]{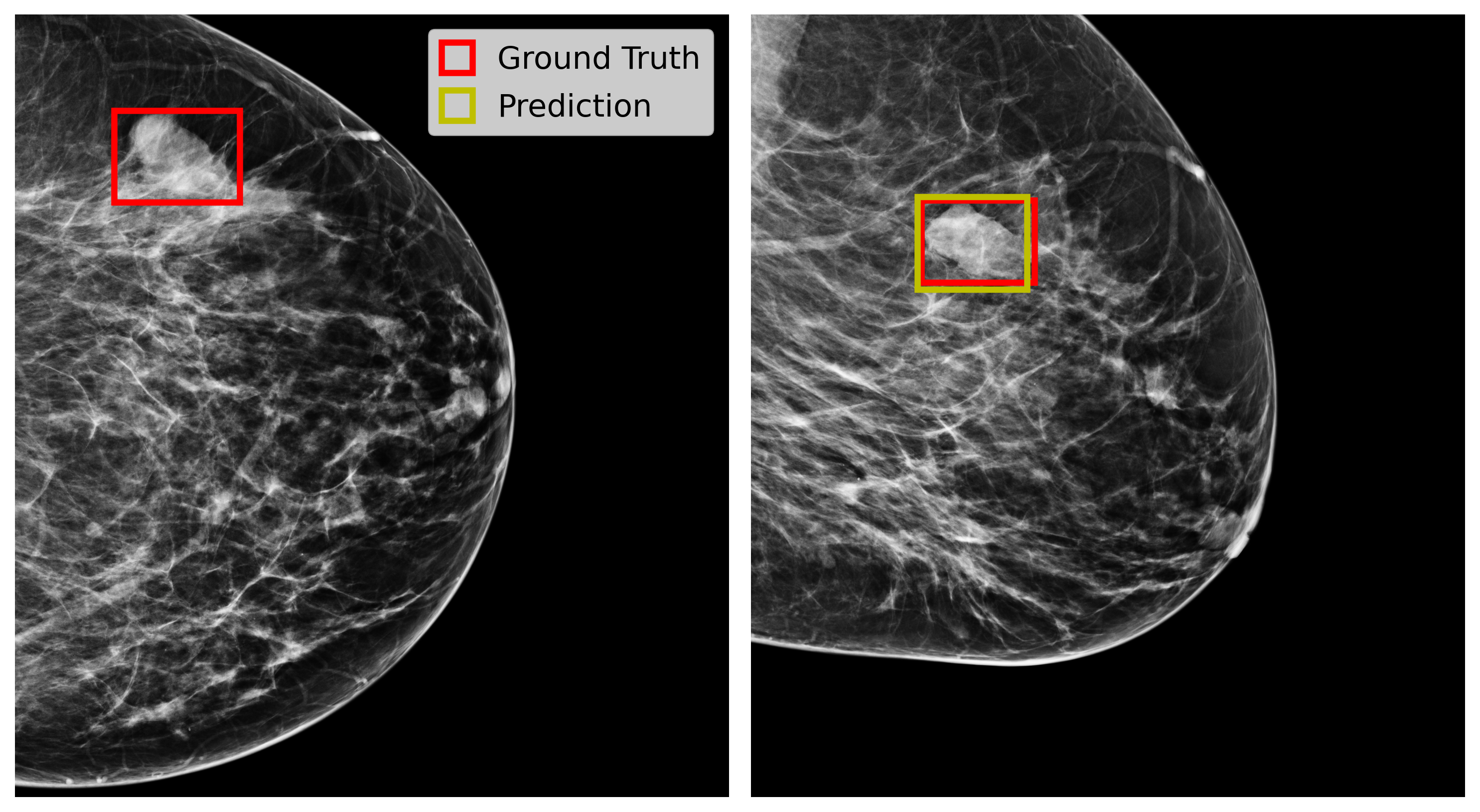}
\end{subfigure} \hfill
\begin{subfigure}[b]{0.47\linewidth}\includegraphics[width=\linewidth]{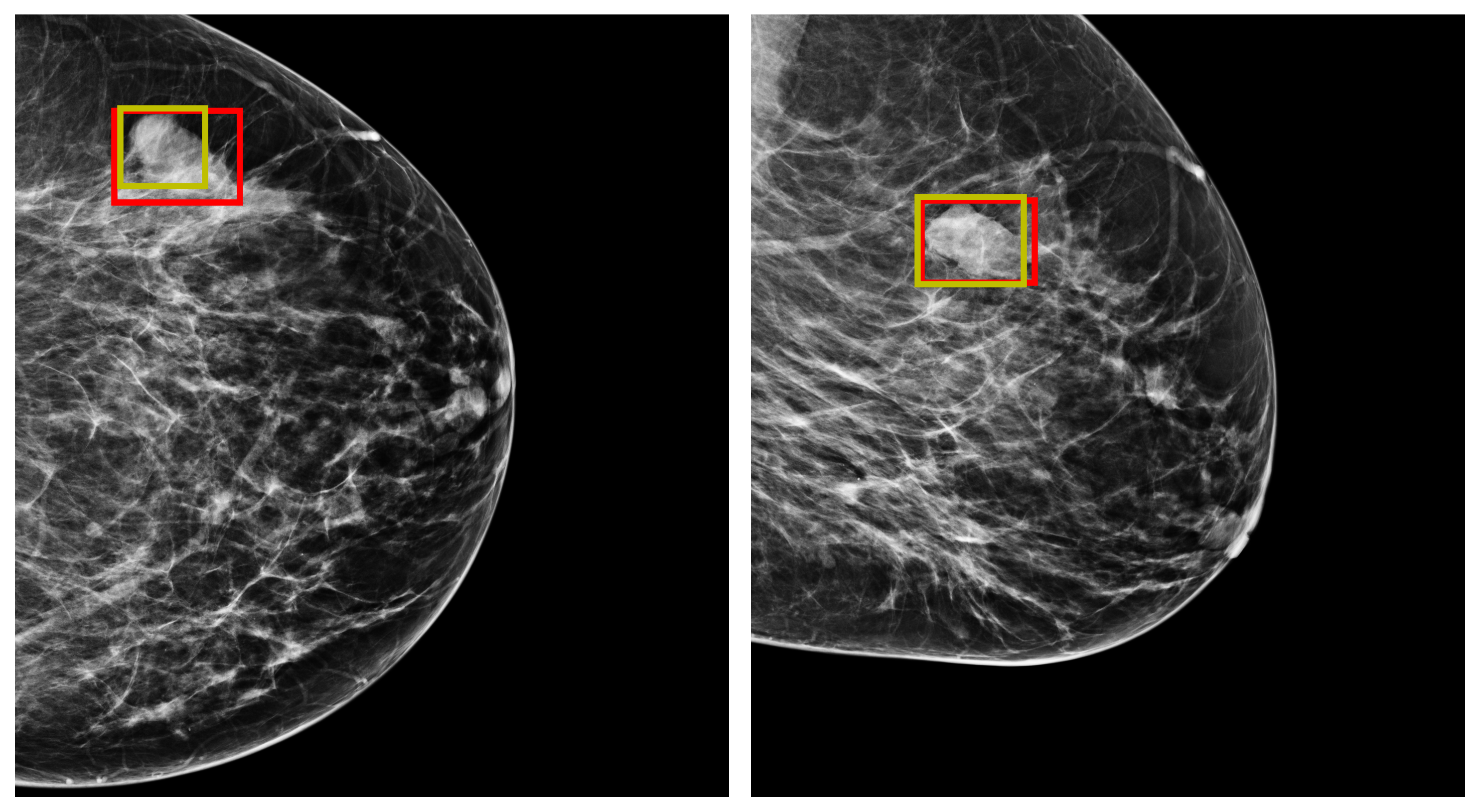}
\end{subfigure}  
\caption{Additional Qualitative Evaluation. \textbf{Left}: without multi-view, the model misses a mass on the CC view even though it was able to detect the mass on the MLO view. \textbf{Right}: with multi-view, M\&M recalls the mass on both views.
}
\end{figure}
\vspace{-1.5cm}
\begin{table}[H]
\setlength{\tabcolsep}{3pt}
    \centering
    \caption{Quantitative classification evaluation with different backbones on 3 datasets. All models are trained using OPTIMAM. M\&M column denotes whether the model was a classifier (-) or M\&M with the row's backbone (\CHECK).}
    \label{tab:backbone_cls}
    \scriptsize
\begin{tabular*}{\linewidth}{@{\extracolsep{\fill}}lcccccc@{}}
\toprule
     \multirow{3}{*}{Backbone} & \multirow{3}{*}{M\&M} & \multicolumn{2}{c}{OPTIMAM} & Inhouse-A & \multicolumn{2}{c}{Inhouse-B} \\
     \cmidrule(lr){3-4} \cmidrule(lr){5-5} \cmidrule(lr){6-7}
     & & Breast    & Exam    & Exam  & Breast     & Exam     \\
     & & AUC & AUC & AUC & AUC & AUC \\
     \midrule
\multirow{2}{*}{GMIC} & - & 0.911 & 0.896 & 0.814 & 0.815 & 0.796 \\
 & \CHECK & 0.920 & 0.900 & 0.835 & 0.843 & 0.822  \\
 \midrule
 \multirow{2}{*}{EfficientNet-B0} & - & 0.940 & 0.922 & 0.787  & 0.850 & 0.826   \\
  & \CHECK & 0.941 & 0.913 & 0.840 & 0.852 & 0.832 \\
  \midrule
 \multirow{2}{*}{PVT-B2-Li} & - & 0.949 & 0.933 & 0.820 & 0.867 & 0.846   \\
  & \CHECK & \textbf{0.960} & \textbf{0.942} & \textbf{0.920} & \textbf{0.910} & \textbf{0.898} \\
\bottomrule
\end{tabular*}
\end{table}
\vspace{-1.2cm}
\begin{table}[H]
    \centering
    \scriptsize
    \setlength{\tabcolsep}{10pt}
    \caption{Quantitative classification evaluation with different backbones on CBIS-DDSM. All models are trained using CBIS-DDSM.}
    \begin{tabular}{@{}ccccc@{}}
    \toprule
    Metric/ Backbone & GMIC & EfficientNet-B0 & ResNet-50 & PVT-B2-Li \\
    \midrule
    Breast AUC & 0.839 & 0.865 & 0.836 & 0.883 \\
    Exam AUC & 0.835 & 0.865 & 0.829 & 0.883 \\
    \bottomrule
    \end{tabular}
    \label{tab:cbis_backbone}
\end{table}

\end{document}